\documentclass[manuscript]{acmart}

\AtBeginDocument{%
  }

\setcopyright{acmcopyright}
\copyrightyear{2023}
\acmYear{2023}
\acmDOI{XXXXXXX.XXXXXXX}

\acmConference[recsys'23]{fashionxrecsys-2023}{Sep. 18,
  2023}{Singapore}

\acmPrice{15.00}
\acmISBN{978-1-4503-XXXX-X/18/06}

\begin{document}

\title{Automated Material Properties Extraction For Enhanced Beauty Product Discovery and Makeup Virtual Try-on}


\author{Fatemeh Taheri Dezaki}
\email{fatimaat@amazon.com}
\author{Himanshu Arora}
\email{hiaror@amazon.com}
\author{Rahul Suresh}
\email{surerahu@amazon.com}
\author{Amin Banitalebi-Dehkordi}
\email{aminbt@amazon.com}
\affiliation{%
  \institution{BeautyTech, Amazon}
  \streetaddress{402 Dunsmuir Street}
  \city{Vancouver}
  \state{British Columbia}
  \country{Canada}
  \postcode{V6B1X4}
}

\renewcommand{\shortauthors}{Taheri Dezaki et al.}

\begin{abstract}


The multitude of makeup products available can make it challenging to find the ideal match for desired attributes. An intelligent approach for product discovery is required to enhance the makeup shopping experience to make it more convenient and satisfying. However, enabling accurate and efficient product discovery requires extracting detailed attributes like color and finish type. Our work introduces an automated pipeline that utilizes multiple customized machine learning models to extract essential material attributes from makeup product images. Our pipeline is versatile and capable of handling various makeup products. To showcase the efficacy of our pipeline, we conduct extensive experiments on eyeshadow products (both single and multi-shade ones), a challenging makeup product known for its diverse range of shapes, colors, and finish types. Furthermore, we demonstrate the applicability of our approach by successfully extending it to other makeup categories like lipstick and foundation, showcasing its adaptability and effectiveness across different beauty products. Additionally, we conduct ablation experiments to demonstrate the superiority of our machine learning pipeline over human labeling methods in terms of reliability. Our proposed method showcases its effectiveness in cross-category product discovery, specifically in recommending makeup products that perfectly match a specified outfit. Lastly, we also demonstrate the application of these material attributes in enabling virtual-try-on experiences which makes makeup shopping experience significantly more engaging.

\end{abstract}



\received{3 August 2023}
\received[accepted]{27 August 2023}

\maketitle

\section{Introduction}

The makeup industry has grown exponentially, with an abundance of products catering to various preferences and skin types. However, this extensive selection poses a challenge for consumers as they navigate through countless options to find the right products that align with their specific requirements. Furthermore, customers often face the frustration of not accurately perceiving the visual attributes of makeup products solely through online catalog images. This lack of certainty necessitates waiting for the product to arrive, leading to a less-than-ideal experience.

Traditionally, e-commerce websites have relied on various sources of information, such as brands, titles, product descriptions, reviews, and recommendations, to guide the purchasing decisions of customers. While these sources can enable ranking of products in a relevant manner but they often fall short in providing a comprehensive understanding of the underlying material properties of the makeup products. Material properties, such as color, finish type and glitter attributes, play a vital role for the customers while making purchasing decisions of makeup products. Customers often have a color in mind and tend to look for similar products within the the same category or across categories. Many times they also are looking particular finish types or glitter shades which aren't available with typical catalog data. Moreover, customers with a dress seeking matching makeup face the challenge of finding cosmetics that complement their attire. Coordinating the makeup with the dress's color, style, and overall aesthetic is their priority to achieve a cohesive and polished look.
Alongside that, Virtual-try on capability~\cite{Soares2019, kips2022real} for makeup products helps makes customers shopping experience more captivating and accurate. Enabling virtual try on also requires extracting material properties from catalog images or textual information available.

\begin{figure*}
  \centering
  \includegraphics[width=0.65\textwidth]{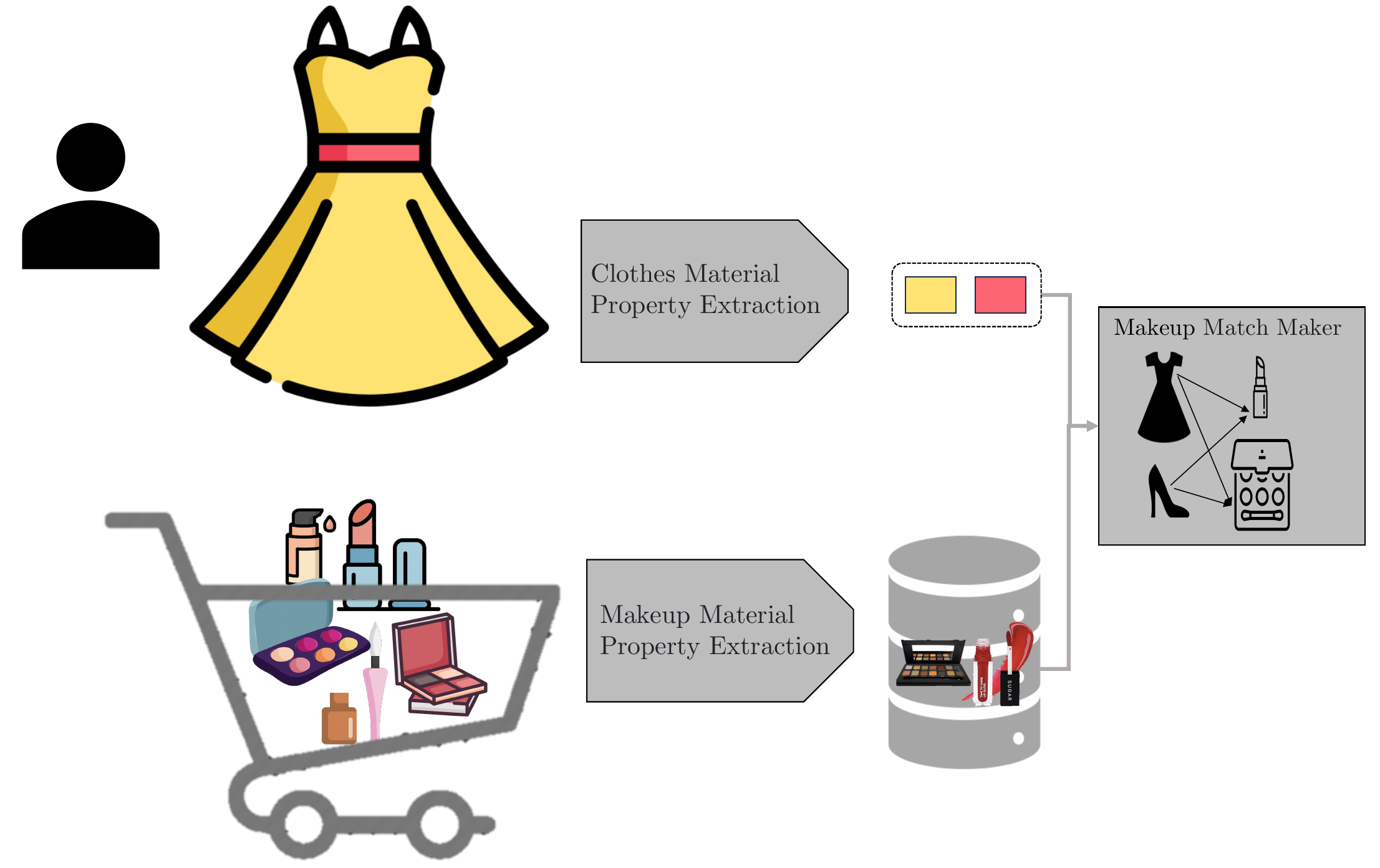}
  \caption{Customer story and proposed solution: A customer has chosen a dress for a specific occasion and seeks matching makeup recommendations based on color and texture attributes.}
   \label{fig:mpe-rec}
\end{figure*}

To tackle these challenges, we propose an innovative approach that centers around extracting material properties from makeup products and leveraging this information to enhance product discovery and recommendations. We propose systematic pipeline to extract key material properties such as color, finish type and glitter color with the help of available product images. Our approach serves a dual purpose: augmenting product discovery capabilities and enabling a virtual try-on experience for makeup products. Using our approach, we can particularly improve recommendations using similar material attribute properties as well as match them to clothing attributes. By employing our approach, we can significantly improve recommendations by considering similar material attribute properties and matching them with clothing attributes. For instance, we can address the problem faced by customers who have a finalized dress for a specific occasion and require matching makeup suggestions based on color and glitter attributes. Our extraction pipeline enables precise and tailored product recommendations for such scenarios (can be seen in Fig~\ref{fig:mpe-rec}). It is important to note that these extracted material properties work in conjunction with the currently used attributes in makeup recommendation. This customization allows for a personalized and tailored recommendation experience based on individual preferences and requirements.

In our work, we propose a novel pipeline specifically designed for extracting makeup attributes. In makeup industry, eyeshadow products exhibit a huge diversity in terms of shade counts, base colors, and finish types, making them a challenging category to analyze. Therefore we consider the variety seen in eyeshadow products in the design of our pipeline, and then showcase the versatility and applicability of our approach to other makeup categories such as lipstick, foundations and etc. Our pipeline has been developed to effectively handle the broad range of shade counts typically found in eyeshadow products, spanning from as low as one to as high as 200. Moreover we design a pipeline for extraction of clothes material properties. Our integrated attribute extraction pipelines for makeup (eyeshadow, lipstick) and clothing enable cross-category product recommendations, enhancing the shopping experience. Our methodology delivers personalized suggestions across diverse makeup and fashion segments, catering to individual preferences and requirements. Our work is closely related to the topic of predicting object attributes and extracting attribute values from images. Prior research has primarily focused on attribute prediction in the fashion domain~\cite{gutierrez2018deep, adhikari2019progressive, guo2019imaterialist}, such as extracting color, texture, and pattern from images of clothing items. In summary, our main contributions are:
\begin{itemize}
   \item We design the material property extraction pipeline with the two-fold aim of enhancing product discovery experience and to facilitate creation of 3D assets required for virtual try-on applications. 
   
   \item We propose an end-to-end learning pipeline for extracting all required material properties for makeup products. 
   
   \item We demonstrate how our extracted attributes can improve relevancy of recommendations for makeup products both from makeup to makeup and outfit to makeup.
   
  \item We perform extensive experiments to show effectiveness and superiority of our learning pipeline and it's applicability for single and multi-shade eyeshadow products
   
   \end{itemize}

 \section{Method}
\label{sec:method}
\begin{figure*}
  \centering
  \includegraphics[width=0.9\textwidth]{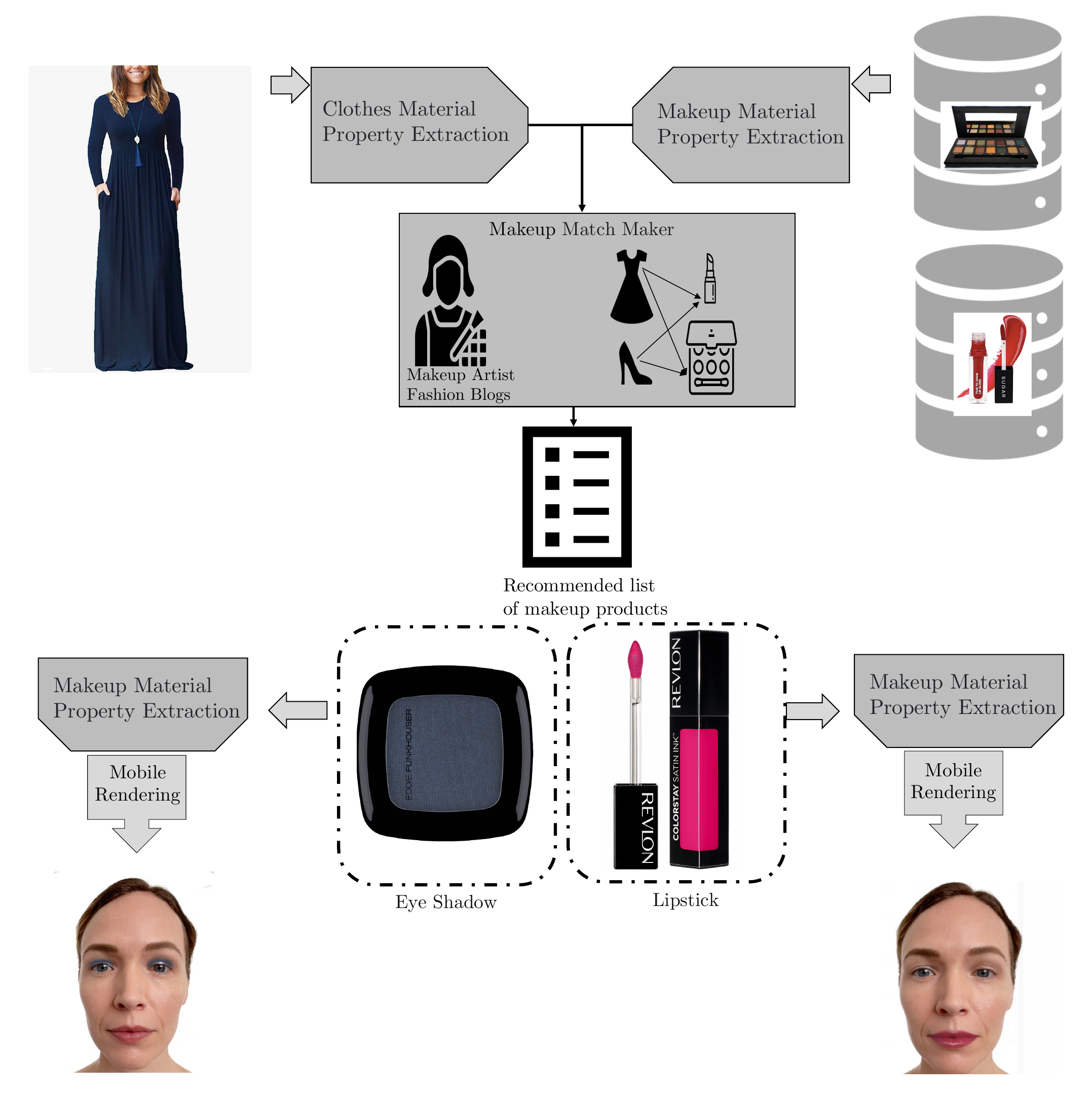}
  \caption{Overview of the proposed pipeline: Three main components include makeup material property extraction, clothes material property extraction, and makeup product match maker.}
   \label{fig:generalpipeline}
\end{figure*}

In this section, we present an  overview of our proposed pipeline (refer to Fig.~\ref{fig:generalpipeline}). The pipeline comprises three main components: makeup material property extraction, clothes material property extraction, and makeup product match maker.

\begin{itemize}
    \item Makeup Material Property Extraction:
This component focuses on extracting material properties related to makeup products. It analyzes various characteristics such as format, finish type, color, and reflectivity. Firstly, we introduce the problem at hand for this main task and subsequently present our proposed solution in the form of an automated material property extraction pipeline. This pipeline aims to address the challenges faced in extracting key material attributes from makeup product images.

 \item Clothes Material Property Extraction:
The second part of our pipeline is dedicated to extracting material properties specifically related to clothing items. By utilizing machine learning algorithms, we can effectively extract and categorize these properties from product images.

 \item Makeup Match Maker:
The final component of our pipeline is responsible for finding the best matching makeup products based on the previous two components that align with the user's desired material properties. It leverages a comprehensive database of makeup products and their corresponding properties. 

\end{itemize}

\subsection{Makeup Material Property Extraction}
An overview of our proposed eyeshadow MPE pipeline can be seen in Fig.~\ref{fig:workflow}. In this section, we first introduce the problem at hand and subsequently present our proposed solution in the form of an automated pipeline. This pipeline aims to address the challenges faced in extracting key material attributes from makeup product images, thereby improving the efficiency and accuracy of the process.

\subsubsection{Problem Formulation}

 Online beauty retailers offer a wide range of eyeshadow products, and several factors contribute to the variety of these products available, including:

\begin{itemize}
\itemsep-0.1em 
    \item \textbf{Brands}: Online beauty retailers offer a wide range of eyeshadow products from different brands, including popular drugstore brands like Maybelline and L'Oreal, as well as high-end brands like Chanel.
    \item \textbf{Colors and finishes}: Cosmetics come in a variety of colors/finishes, like matte, shimmer, metallic and glitter. Online beauty retailers offer a range of color options, from natural tones to bold and bright colors.
    \item \textbf{Formulations}: Eyeshadows are available in different formulations, such as powder, cream, stick and liquid. Each formulation offers a unique finish and texture, allowing customers to choose the one that best suits their needs.
    \item \textbf{Packaging}: Eyeshadow palettes are available in different packaging, such as individual pans, duos, quads, and larger palettes with multiple shades. This allows customers to choose the packaging that best suits their preferences and needs.
\end{itemize}

The key elements affecting the look of a beauty product applied on user’s face includes application style, finish type, color, pigmentation/coverage, etc.  From a user study, we identified the common application styles and concluded to have the material properties required for eyeshadow rendering as:
\begin{figure}
  \centering
  \includegraphics[width=0.75\linewidth]{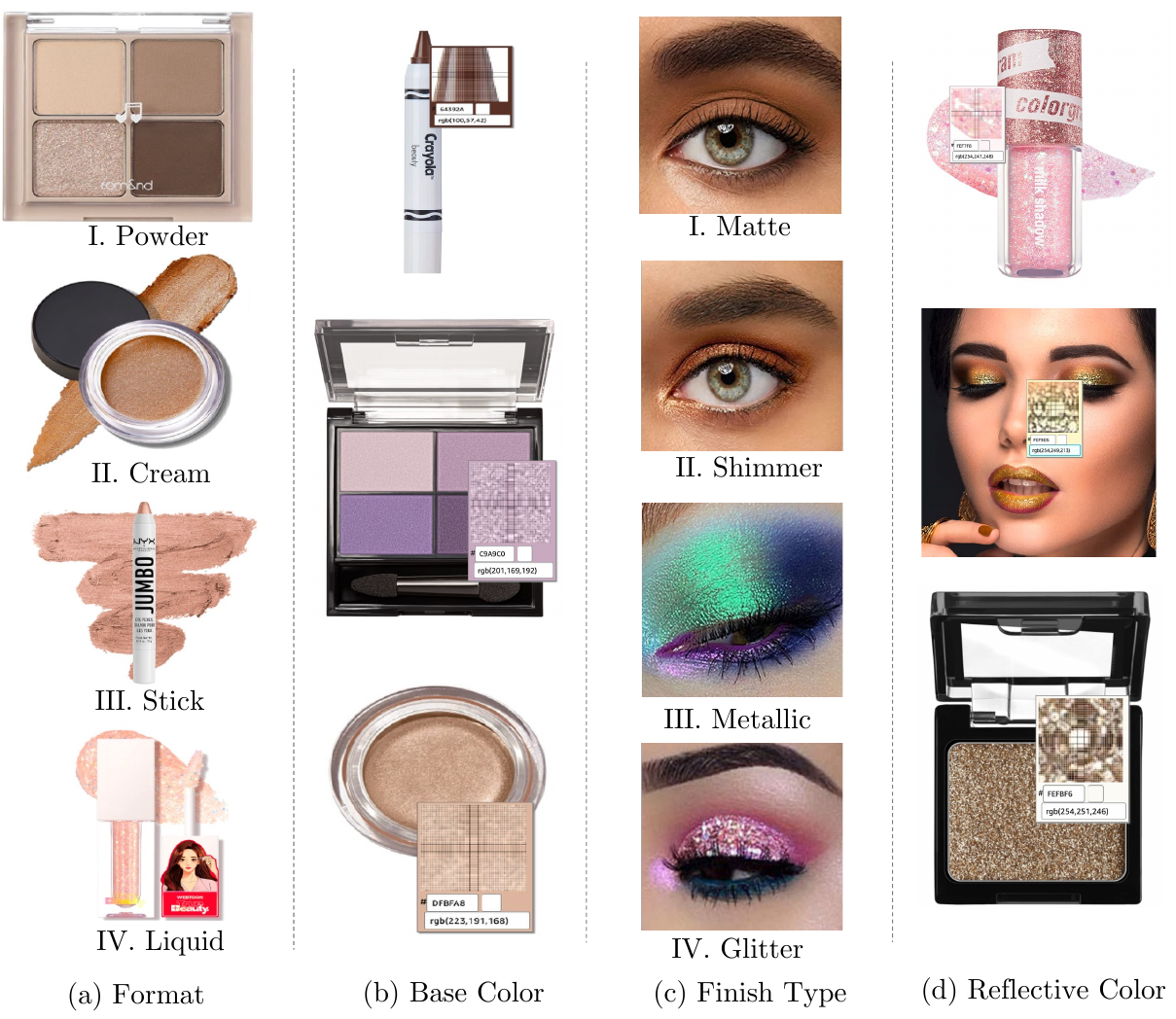}
  \caption{Eyeshadow material properties}
   \label{fig:eyeshadow_properties}
\end{figure}
\begin{itemize}
\itemsep-0.1em 
    \item \textbf{Format}: powder, cream, stick and liquid. Please see Figure~\ref{fig:eyeshadow_properties}.(a) for more illustration.
    \item \textbf{Base color} (RGB hex code). Please see Figure~\ref{fig:eyeshadow_properties}.(b) for more illustration.
    \item \textbf{Finish type} (one of): matte, shimmer, matallic, glitter. Please see Figure~\ref{fig:eyeshadow_properties}.(c) for more illustration.
   \item \textbf{Reflective color} (if any)(rgb hex code). Please see Figure~\ref{fig:eyeshadow_properties}.(d) for more illustration. 
   \end{itemize}

 In order to extract material properties of eyeshadow products from a detail page in an efficient and scalable manner, we propose a machine learning pipeline including several trained models specialized for each task in the MPE process.

\subsubsection{Overall Proposed Model Architecture}

\begin{figure*}[h]
  \centering
  \includegraphics[width=0.9\textwidth]{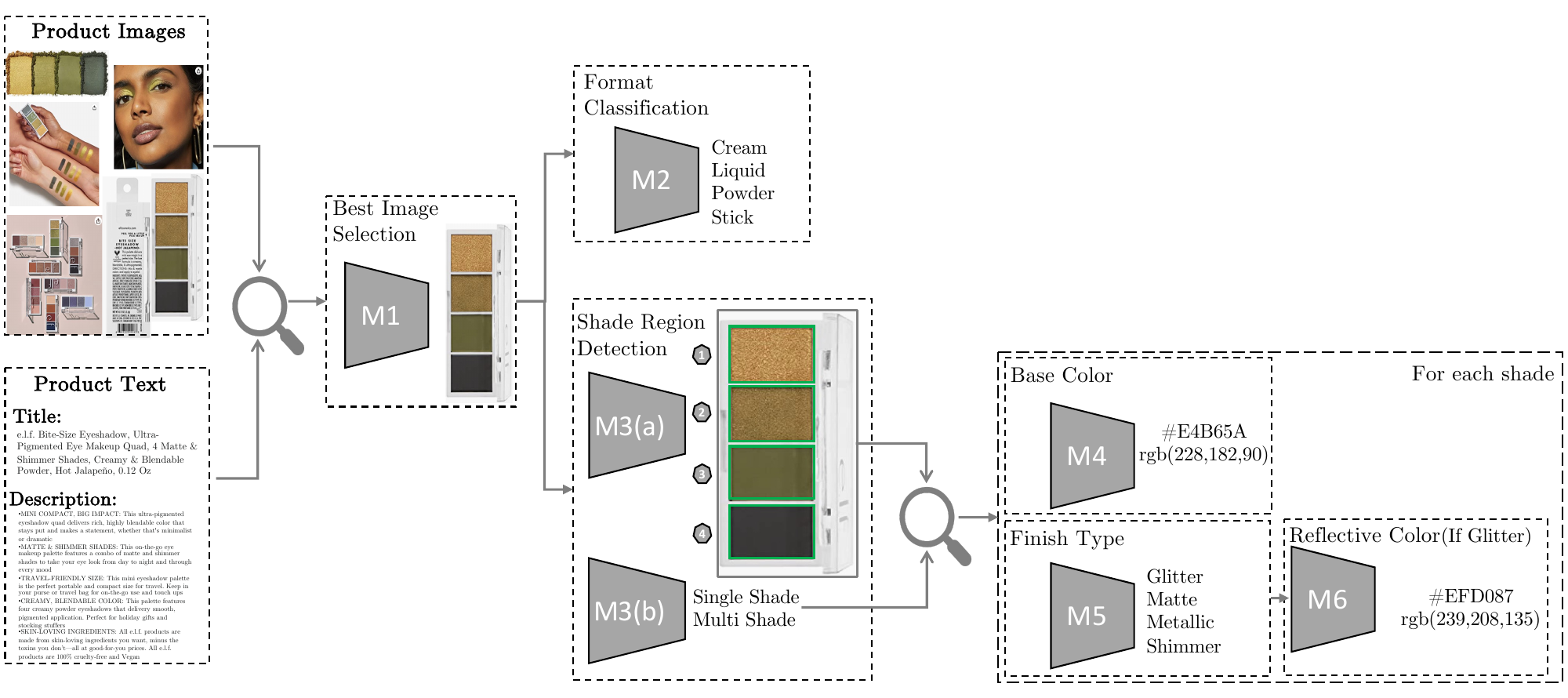}
  \caption{Overall pipeline architecture}
   \label{fig:workflow}
\end{figure*}

Our MPE pipeline consists of six models each responsible for extracting a specific property of a given eyeshadow product. A visual description of the entire pipeline of eyeshadow MPE is shown in Fig.~\ref{fig:workflow} and a detailed explanation of each component is provided in the following subsections.  Given any eyeshadow product we should be able to extract the material properties required for eyeshadow rendering. The input to this pipeline is the product images and text descriptions of that eyeshadow product. At first, we have a simple filtering mechanism to check if the given product can be fed to the pipeline, for example some products are eyeshadow organizers/cases (they hold eyeshadow, but are not eyeshadow products) or the product shows makeup kits or bundles (there are more than just eyeshadow in the product) or they are multi-chrome eye or florescent/neon type eyeshadow which are not supported with the current rendering approach. For such filter we parse the product title and look for specific key words (and their variation), like  \textit{multichrome, iridescence, makeup kit, makeup organizer, ...} . After this step, images are the only input for inferring the material properties of eyeshadow products. 

To do so, as a next task we need to select the best representative image among, $I^p_{best}$, the varied images present for the product on the online beauty retailers' web pages, which is representative enough for MPE.  The selected image is used as the input for the rest of the pipeline. First, it passes through a classification model to identify the format of the eyeshadow product, followed by a shade region detection model that detects the number of shades in the given product $p$ and their locations in the picture. To ensure accuracy, we trained a binary classification model to predict whether the product $p$ includes a single shade or more than one, which helps filter out incorrectly predicted regions from the previous model. For each detected shade, the base color and finish type are predicted using the region detected in the previous step. Finally, if the detected finish type is  \textit{glitter}, the reflective color is identified for rendering the sparkles using the last model in the pipeline.

\subsubsection{Best Image Selection}
In the online beauty retailers webpages, the catalog images have some orders to be displayed, but most often the orders don't imply any semantic information and also most sellers don't follow the prescribed order of image types.  To fulfill our task, we require an image that displays all shades clearly, without being obstructed by any text or packaging. We perform this task by training a model which learns to choose the preferred image in a binary manner i.e selects the 'preferred' image in a pair. We individually input two images into ResNet50\cite{he2016deep} based shared encoder, and then merge their features together to make a prediction about the preferred image. In the inference process, the trained model compares other images to the MAIN i.e first image in the online beauty retailers' webpages image as a reference. If the model predicts any image as being preferred over the MAIN image, we select it for downstream tasks. In the case where multiple images are predicted as preferred over the MAIN image, we choose the one with the highest confidence score to be used for the downstream tasks.


\subsubsection{Shade region detection}
\label{subsec:shade_detection}
\begin{figure}
  \centering
  \includegraphics[width=0.8\textwidth]{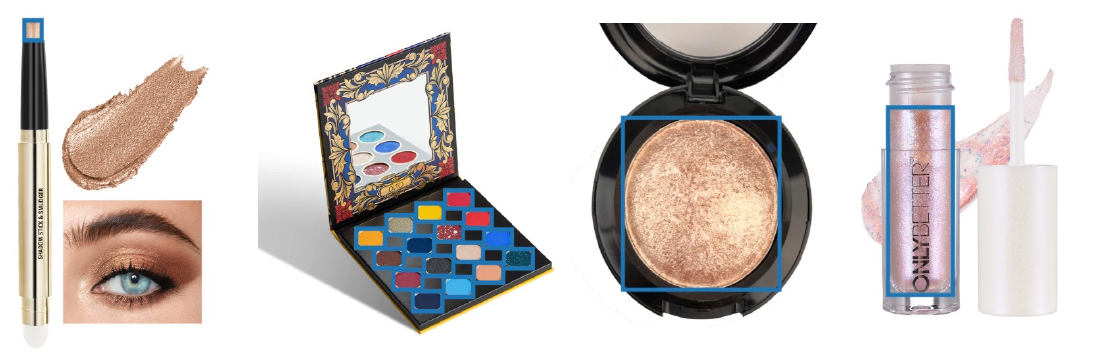}
  \caption{Bounding boxes overlayed on different eyeshadow products with blue color.}
   \label{fig:bbox}
\end{figure}

Given the best image selected, $I^p_{best}$, we would like to locate all the existing shades with a bounding box: center point coordinates, width and height, $\ (x_i, y-i, w_i, h_i), \ i \in \{1,... ,s\}$. To do so, a popular object detection model , YOLOV5~\cite{yolov5}, was trained on in-house dataset. In Fig.~\ref{fig:bbox} some sample eyeshadow products with the corresponding bounding boxes to show each shade location is shown. YOLOV5 utilizes a single-stage object detection approach, meaning it processes the entire image only once, which allows it to perform real-time object detection on high-resolution images. YOLOV5 is based on a deep convolutional neural network that consists of a backbone, neck, and head. The backbone is responsible for extracting features from the input image, while the neck and head perform object detection and classification tasks. The backbone of YOLOV5 uses a variant of the CSPNet (Cross-Stage Partial Network) architecture~\cite{wang2020cspnet}, which is designed to improve the learning capability of convolutional neural networks by facilitating cross-stage feature aggregation. This architecture allows YOLOV5 to process input images efficiently while preserving spatial resolution. The neck of YOLOV5 is composed of a series of convolutional layers that are responsible for fusing features from different levels of the network. This process is called feature pyramid fusion, and it allows YOLOV5 to detect objects of different sizes and scales. The head of YOLOV5 consists of a set of convolutional layers that predict bounding boxes and class probabilities for detected objects. The output of the head is a set of bounding boxes with confidence scores and class probabilities, which are then used to identify and localize objects in the input image.
To ensure the accuracy of our model, we trained a binary classification algorithm for eyeshadow products with a single shade. This allows us to predict whether the product $p$ contains one or more shades, which in turn helps to filter out incorrectly predicted regions from the YOLOV5 model. A Resnet50 model was fine-tuned for this binary classification task. For instance, if the image is classified as a single-shade product, but the YOLOv5 model detects multiple regions, we can select the region with the highest confidence score and pass it on to the subsequent stages of the pipeline.

\subsubsection{Format-finishtype classification}
The pipeline includes two classification models: one for detecting the format of the eyeshadow product based on the best image, $I_{best}$, and another for detecting the finish type of each eyeshadow shade, $s$, based on cropped regions of the same image, $I{^s_{best}}$. Various model architectures were experimented with for each of the classification tasks, and the best architecture was selected based on factors such as data size and final accuracy. 

Out of all the models that were tested for format classification, EfficientNet-B3~\cite{tan2019efficientnet} exhibited the best performance.  It is part of the EfficientNet family of models, which are designed to achieve state-of-the-art performance on a range of computer vision tasks while minimizing the number of parameters and computations required. EfficientNet-B3 has a relatively large number of layers, with a total of 76 convolutional layers, and it uses a combination of depthwise separable convolutions, squeeze-and-excitation blocks, and a compound scaling method to achieve high accuracy and efficiency. One of the key innovations of the EfficientNet architecture is the use of a compound scaling method, which scales the width, depth, and resolution of the model simultaneously. This allows the model to achieve high accuracy while using fewer parameters and computations than other models of comparable size.

Among the different models we experimented with, Resnet50~\cite{he2016deep} with 4 classes performed the best in the finish type classification task. The input to this model is the cropped region detected in the previous step. As the finish type classification task is highly imbalanced with respect to the 4 classes, with a greater number of Matte and Shimmer finishes compared to Metallic and Glitter finishes, we employed various techniques to balance the classes, including data augmentation and oversampling of the classes with lower populations.

\subsubsection{Basecolor-reflective color regression}
To detect the base and reflective colors (if any) of the shade region detected in the previous section~\ref{subsec:shade_detection}, three values for normalized RGB channels are predicted. Resnet50 was demonstrated to be a suitable choice as the backbone for these models. The base color of available eyeshadow products on in our dataset exhibits a high degree of variability, ranging from the darkest to the brightest values in each of the RGB channels. On the other hand, reflective colors mostly tend to have brighter values. Based on this observation, we utilized the raw RGB values for detecting the base color, whereas the reflective color is scaled after being predicted. For instance, if we take into account the predicted values for the R channel, we apply an additional scaling using the equation $aR + b$, where $a=0.4$ and $b=0.6$, to prevent the model from producing trivial outputs regardless of the input it receives.

\subsection{Clothes Material Property Extraction}
\begin{figure}
  \centering
  \includegraphics[width=0.8\textwidth]{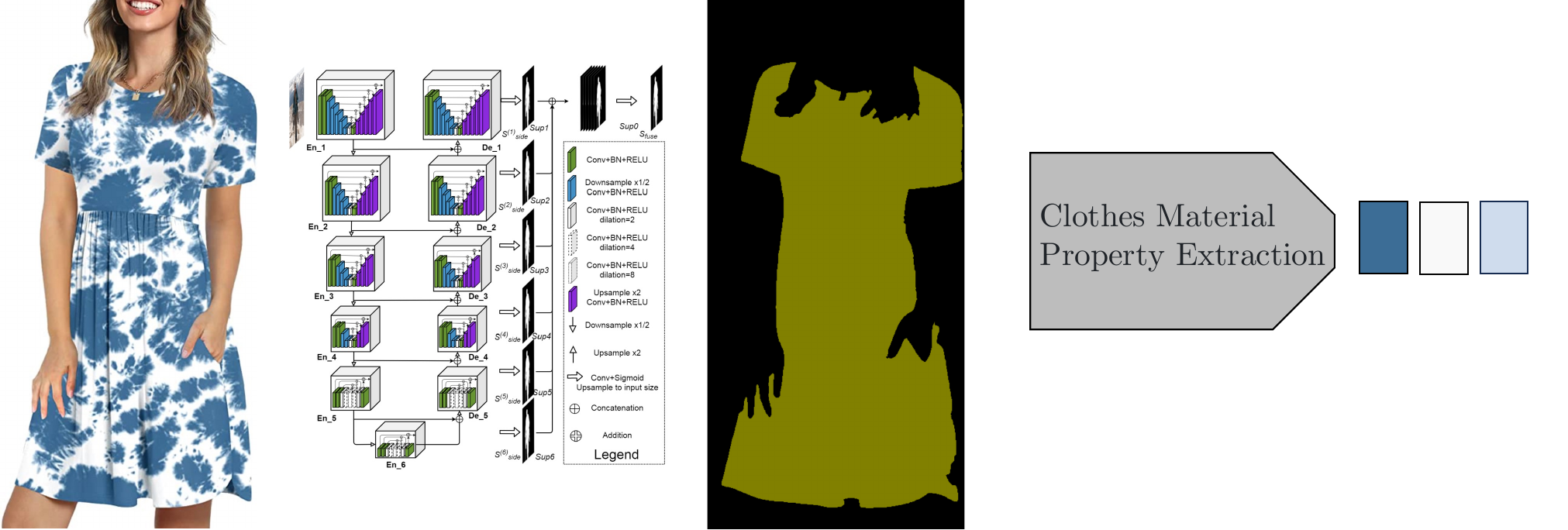}
  \caption{Clothes color extraction from identified region using U2NET segmentation.}
   \label{fig:u2net_color}
\end{figure}

The clothes material property extraction process is comprised of two key components (refer to Fig~\ref{fig:u2net_color}): a segmentation model and an attribute prediction model. For the clothes segmentation model, we employed U2NET~\cite{QIN2020107404}, a deep learning model renowned for its effectiveness. Unlike the conventional salient object detection task where a single-channel output is obtained, our modified U2NET produces four channels. Each channel represents a specific clothing region, namely upper body cloth, lower body cloth, fully body cloth, and background. This model demonstrates excellent performance across various backgrounds and poses.

 By leveraging U2NET's multi-channel output and employing appropriate loss functions, we could precisely identify and classify different clothing regions in an image. This allows for more precise analysis and prediction of clothing attributes in subsequent stages of the pipeline.

Once the desired region is identified, we proceed to extract the top three to five colors present in the clothes. To accomplish this, we employ the k-means clustering algorithm, which enables us to identify the dominant colors within the clothing region. These extracted colors play a crucial role in determining the best makeup product that complements the chosen clothing.

\subsection{Makeup Match Maker}
We leverage the extracted color and finish types to search for relevant products in our catalog, utilizing the extracted material attributes. We seek input from makeup professionals who suggest suitable makeup i.e.lipstick and eyeshadow options that complement the dresses. By incorporating material property extraction (MPE) attributes for makeup products, we strive to deliver recommendations that align with human preferences. Although we currently employ a manual matching strategy due to the subjective nature of preferences, automated rules can be established to scale the recommendation system.

\section{Experiments}

\subsection{Datasets}
Experiments were conducted on a collection of makeup/clothing product images crawled from online shopping webpages. The eyeshadow data utilized in the experiments comprised of 3700  products that appear to be top sold products on shopping websites.
The eyeshadow products have varying numbers of images to display, with an average of 5 images per product. The number of images can range from a minimum of one to a maximum of twenty. The data was divided into two separate sets, a training set and a validation set, using a mutually exclusive split with a ratio of 70-30\% respectively.

\begin{figure*}
  \centering
  \includegraphics[width=0.8\textwidth]{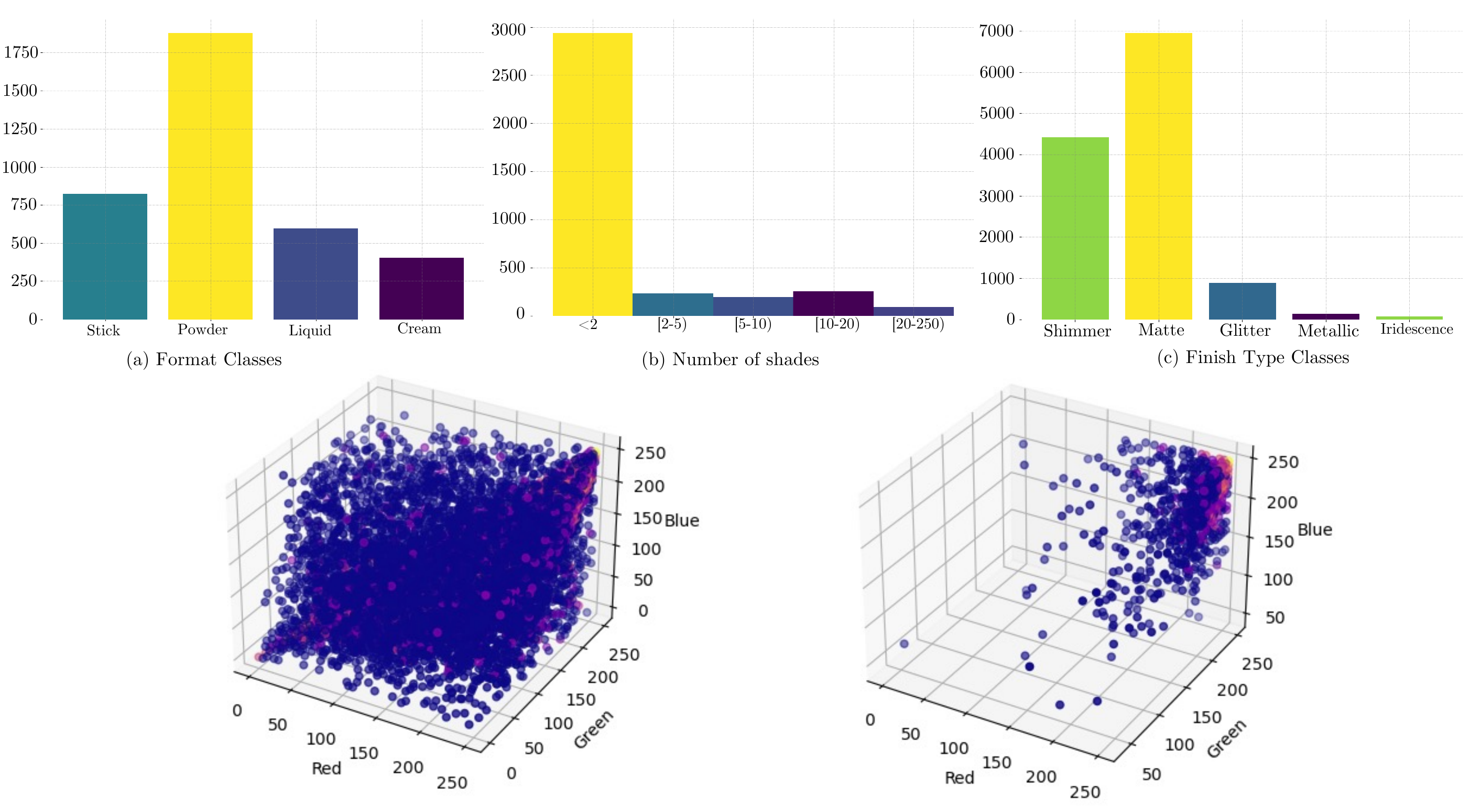}
  \caption{Data visualization}
   \label{fig:dataset}
\end{figure*}

Fig~\ref{fig:dataset} provides a comprehensive view of the eyeshadow dataset. The data shows that "Powder" is the most common format followed by stick, liquid, and cream, respectively. In addition, Fig~\ref{fig:dataset}.(b) reveals that most eyeshadow products have a single shade, although some have as many as 250 shades in one product. Matte is the most prevalent finish type, followed by Shimmer, with Glitter and Metallic being less common. Iridescent shades, which feature multiple colors in one shade, were filtered based on the text in detail page.

Furthermore, the 3D map of base color and reflective color in RGB space highlights the color distribution in Fig~\ref{fig:dataset}.(d,e). While there are many variations in the base color of eyeshadow products, the reflective color is mostly concentrated in one corner of RGB space, with large values in all three channels of R, G, and B.

\subsection{Results}

To evaluate the performance of each ML model in our pipeline we used different evaluation metrics and report them in Table~\ref{tab:performance}. These metrics include

\begin{itemize}
\itemsep-0.1em 
    \item \textbf{deltaE} \cite{ortiz2016evaluating} is a metric to quantify distance between two colors. It is computed as euclidean distance between two points in CIELAB colorspace.
    \item \textbf{mAP} \cite{felzenszwalb2009object} is mean average precision use to quantify how many objects are correctly localized in the prediction. We show mAP for objects with IoU between $0.5$ to $0.95$.
    \item \textbf{F1-Score} is computed by  calculating harmonic mean of precision and recall of classification model.
    \item \textbf{Selection Accuracy} is the count of times ML model selection matches human selection.
\end{itemize}

\begin{table*}[t]
\caption{End-to-end models performance for all individual tasks in the pipeline.}
\centering
\resizebox{0.9\linewidth}{!}{%
\begin{tabular}{lllllllll}
\toprule
                & \multicolumn{3}{l}{}         & \multicolumn{2}{c}{Single Shade} & \multicolumn{2}{c}{Multi Shade} &            \\ 
                \cmidrule(r){5-6} \cmidrule(r){7-8}
                & M1 (Acc$\uparrow$) & M2(f1-score$\uparrow$) & M3(mAP$\uparrow$) & M4(deltaE	$\downarrow$)       & M5(F1-Score$\uparrow$)       & M4(deltaE	$\downarrow$)       & M5(F1-Score$\uparrow$)      & M6(deltaE	$\downarrow$) \\
                & Best Image & Format & Shade Region  & Base Color  & Finish type  &  Base Color  & Finish type  & Reflective Color \\ 
                
                & Selection &  Classification & Detection &  Identification &  Classification &  Identification & Classification & Identification \\ 
                \midrule
NO GT           & 92.22    & 0.90   & 0.93    & 6.35             & 0.84         & 4.86             & 0.89        & 6.18       \\ 
GT M1           & NA       & 0.91   & 0.94    & 6.35             & 0.84         & 4.76             & 0.89        & 6.18       \\ 
GT M1 + M3      & NA       & 0.91   & NA      & 5.89             & 0.87         & 4.83             & 0.9        & 5.85       \\ 
GT M1 + M3 + M5 & NA       & 0.91   & NA      & 5.89             & NA           & 4.83             & NA          & 6.28       \\ 
\bottomrule
\end{tabular}
}

\label{tab:performance}
\end{table*}

Table~\ref{tab:performance} displays the performance of our models under multiple scenarios. In the first row, we
use all the models’ predictions to evaluate end-to-end impact. We also show separate analysis of
single shade and multi shade for both finish type prediction(M5) and base color(M4) as they are the
key end points. In the second row, we do not use Best Image Selection Model (M1) predictions and
instead the GT data for image selection has been used. We observe that there’s a slight difference
which is propagated due to error in M1. For example M3 goes from 0.93 mAP to 0.94. Similarly,
in following row we substitute both M1 and M3 (Shade Detection) with ground truth to observe
error getting carry forward. In the last row, we use ground-truth for finish type prediction so we can
observe the performance of M6. M6 delta E increases because more shades are Glitter in Ground
Truth as compared to using M5 prediction model. We show that the deltaE for base color identification is ~6 for single shade and ~5 in multi-shade which is pretty good for color accuracy and falls within the range where it becomes difficult for an human to differentiate. 

\begin{figure*}
  \centering
  \includegraphics[width=0.85\textwidth]{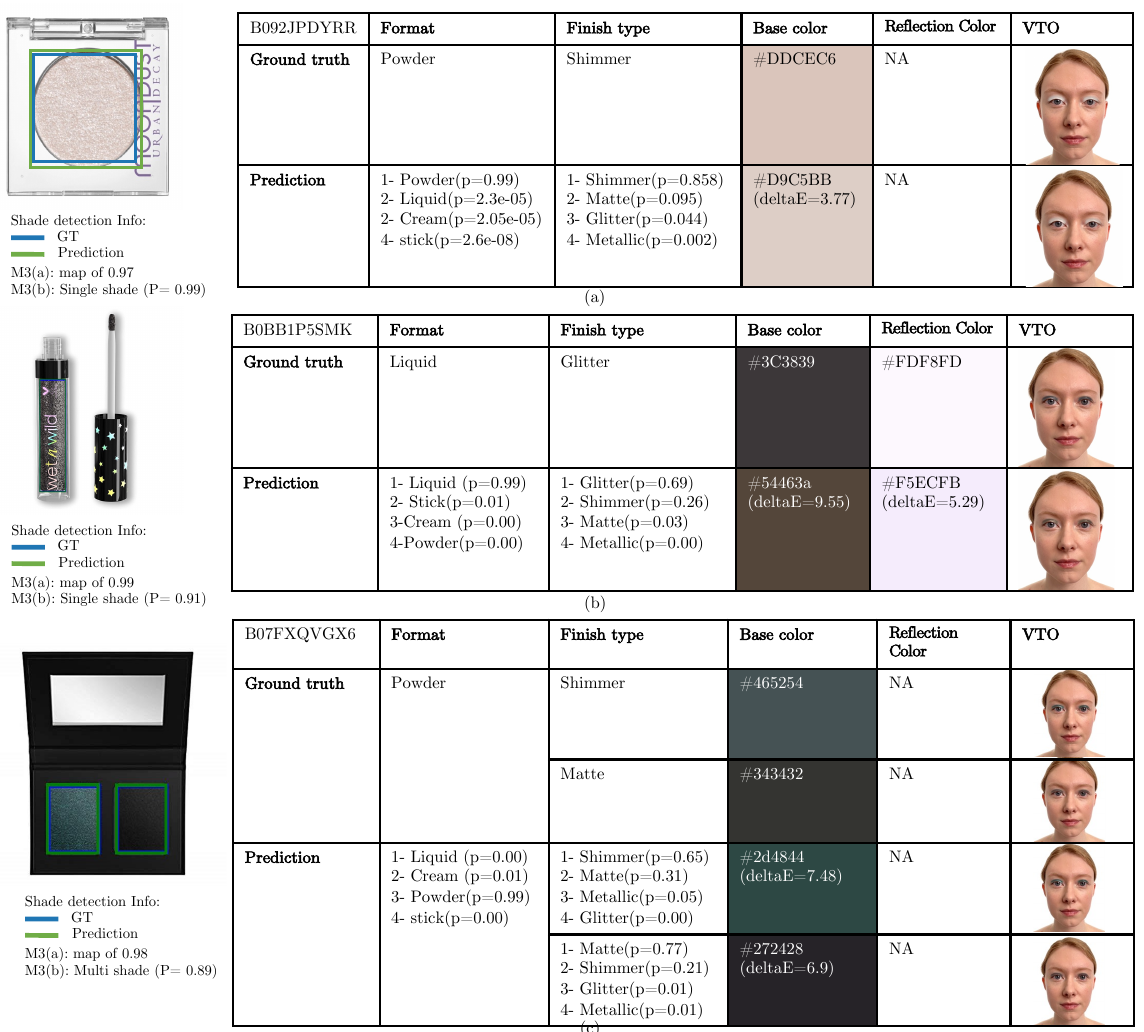}
  \caption{End to end pipeline predictions and VTO on static model images}
   \label{fig:end2end_visualization}
\end{figure*}

\begin{figure}[tbh]
  \centering
  \includegraphics[width=0.95\linewidth]{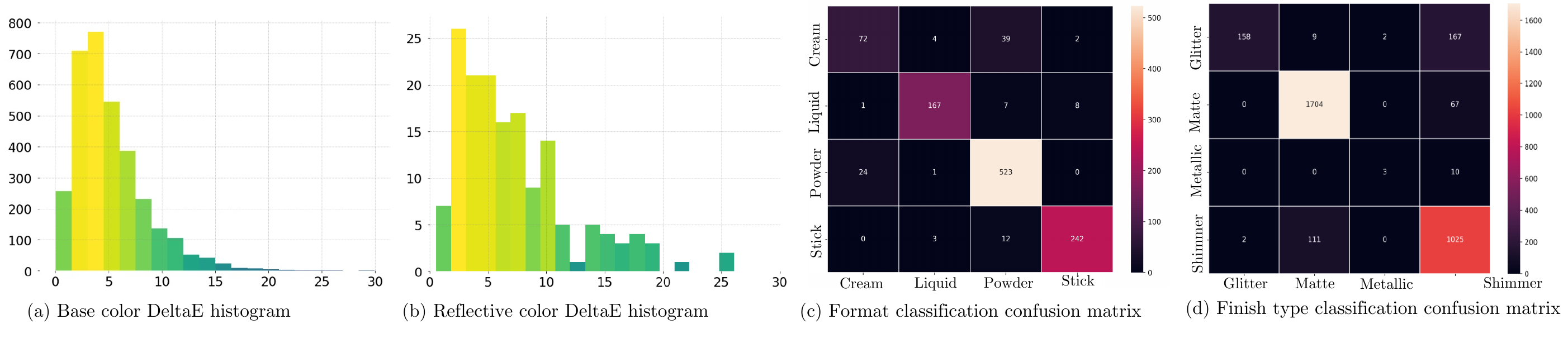}
  \caption{Performance visualization of DeltaE distance histograms of the base color and reflective color, and confusion matrices for format and finish type classification models}
   \label{fig:individual}
\end{figure}

We demonstrate various types of eyeshadow products in Fig~\ref{fig:end2end_visualization} and compare the ground truth annotations with the predictions generated by our individual pipeline models. In the last column, we additionally compare the appearance of the rendered eyeshadow product on static images of a human model. As it can be seen our proposed pipeline outputs are comparable to or even superior to that of human annotations. It can be further observed that despite minor differences with respect to Ground Truth in intermediate models, the final virtual try-on experiences are very close to as with ground-truth data. This demonstrates the robustness of our pipeline.  Additional visualizations with higher resolution can be found in the supplementary material section.

In Fig~\ref{fig:individual}.(a,b), we present histograms of deltaE scores between the ground truth and our color models' predictions. The data shows that the majority of the deltaE scores for both the base color and reflection color models fall below 10, which is an acceptable range. More specifically, our analysis shows that 30\% of the base color model predictions have a deltaE score of less than or equal to 3, which is imperceptible to most human eyes, while 65\% have a deltaE score within 3-12, which is distinguishable but still acceptable to human eyes.

\begin{figure}[tbh]
  \centering
  \includegraphics[width=0.8\linewidth]{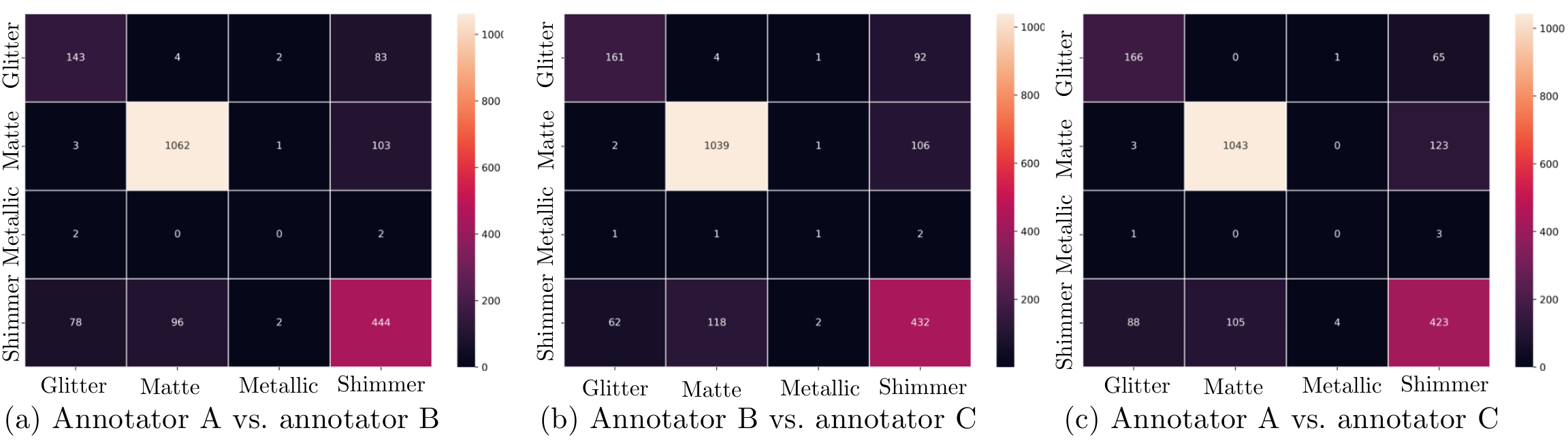}
  \caption{Annotators agreement for finish type classification in our dataset}
   \label{fig:annotator}
\end{figure}

Fig~\ref{fig:individual}(c,d) shows the confusion matrix for our finish type and format classifications. While the finish type model performs well for most classes, it struggles with differentiating between \textit{glitter} and \textit{shimmer} finishes. We confirmed that this confusion is difficult for human annotators as well by studying their confusion for finish type classification labels in Fig~\ref{fig:annotator}. Despite this confusion, our qualitative results demonstrate that these two classes are perceptually similar, and the final VTO rendering is nearly identical if the base color is predicted accurately.

In one of our experiments we evaluated the performance of base color and finish type ML models with respect to product images from different makeup brands. The experiment showed that both models performed better than average for key brands with top eyeshadow products. This is due to more uniform image quality from these larger brands. The experiment also helped identify brands with unsuitable images for MPE tasks, such as Revlon, showing a lower finish type F1-score than the average. More detailed results of this experiment can be seen in Supplementary Material section.

\subsection{Ablation Studies}

To assess the inter-human consistency, we gathered a group of individuals to perform the same task of annotating eyeshadow material properties. We then calculated the level of agreement between the three annotators for two main eyeshadow properties: base color and finish type. By comparing the performance of the ML model with the inter-annotator agreement among the three annotators, we can gauge the level of agreement between the annotators and determine if the model is performing better or worse than humans on average.

As for assessing base color model predictions and measuring the consistency of our model and that of human annotators, we compute the discrepancy among human annotators for any shade $s$ as:
\begin{equation*}
\resizebox{.45\hsize}{!}{$d_{HC} = dist(A1_s,A2_s) + dist(A1_s,A3_s) + dist(A2_s,A3_s)$}
\end{equation*}
where $dist(x,y)$ refers to the deltaE distance between two colors and $A1_s,A2_s,A3_s$ refers to color annotated by each of three annotators for shade $s$. Similarly, we compute the discrepancy for each shade our ML model using
 \begin{equation*}
\resizebox{.45\hsize}{!}{$d_{ML} = dist(A1_s,ML_s) + dist(A2_s,ML_s) + dist(A3_s,ML_s)$}
\end{equation*}
where ${ML}_s$ refers to predicted base color by ML model for shade $s$. We show the mean and variance in the table \ref{human_consistency}. Assuming the mean of human annotations as ground-truth, we observed that the ML model's distance to the ground-truth is comparable to the sum of differences among human annotations. However, the variance in the ML model's predictions is significantly lower, indicating that the model is less likely to deviate from the ground-truth. This suggests that the ML model provides consistent and less subjective ratings compared to human annotators.
 
\begin{table}[htb]
\caption{Human Consistency Analysis}
\label{human_consistency}
\centering
\footnotesize
\begin{tabular}{ccccccc}
\toprule
          & \multicolumn{2}{c}{Single-Shade} & \multicolumn{2}{c}{Multi-Shade} & \multicolumn{2}{c}{All-Shades} \\ \cmidrule(r){2-3} \cmidrule(r){4-5} \cmidrule(r){6-7}
          & Mean          & Variance         & Mean          & Variance        & Mean         & Variance        \\ \midrule
$d_{HC}$ & 6.31         & 62.48           & 11.82         & 112.84         & 6.42        & 64.10          \\ 
$d_{ML}$ & 6.36         & 34.43           & 10.2         & 59.95          & 6.44        & 35.24           \\ 
\bottomrule
\end{tabular}
\end{table}

As for analyzing performance of our classification tasks, we also compare the F1-score of ML model predictions with the Fleiss Kappa coefficient computed on human annotations. Fleiss Kappa coefficient is a commonly used measure of inter-rater agreement for categorical items ~\cite{fleiss1971measuring, he2021annotator}. As can be seen in Table~\ref{fleisskappa}, we observe that our f1-score is similar to the Fleiss Kappa coefficient for the format classification task, and significantly better for finish type prediction. This indicates that our ML model is equally or more reliable than human annotations for both classification tasks.

\begin{table}[htb]
\caption{Fleiss Kappa Coefficient and F1-score for Finish Type and Format Classification}
\label{fleisskappa}
\centering
\footnotesize
\begin{tabular}{lll}
\toprule
                       & Fleiss-Kappa & f1-score \\
                       \cmidrule(r){2-3}
Format Prediction      & 0.91         & 0.902    \\
Finish Type Prediction & 0.66         & 0.870    \\
\bottomrule
\end{tabular}

\end{table}

\begin{figure}[htb]
  \centering
  \includegraphics[width=0.9\linewidth]{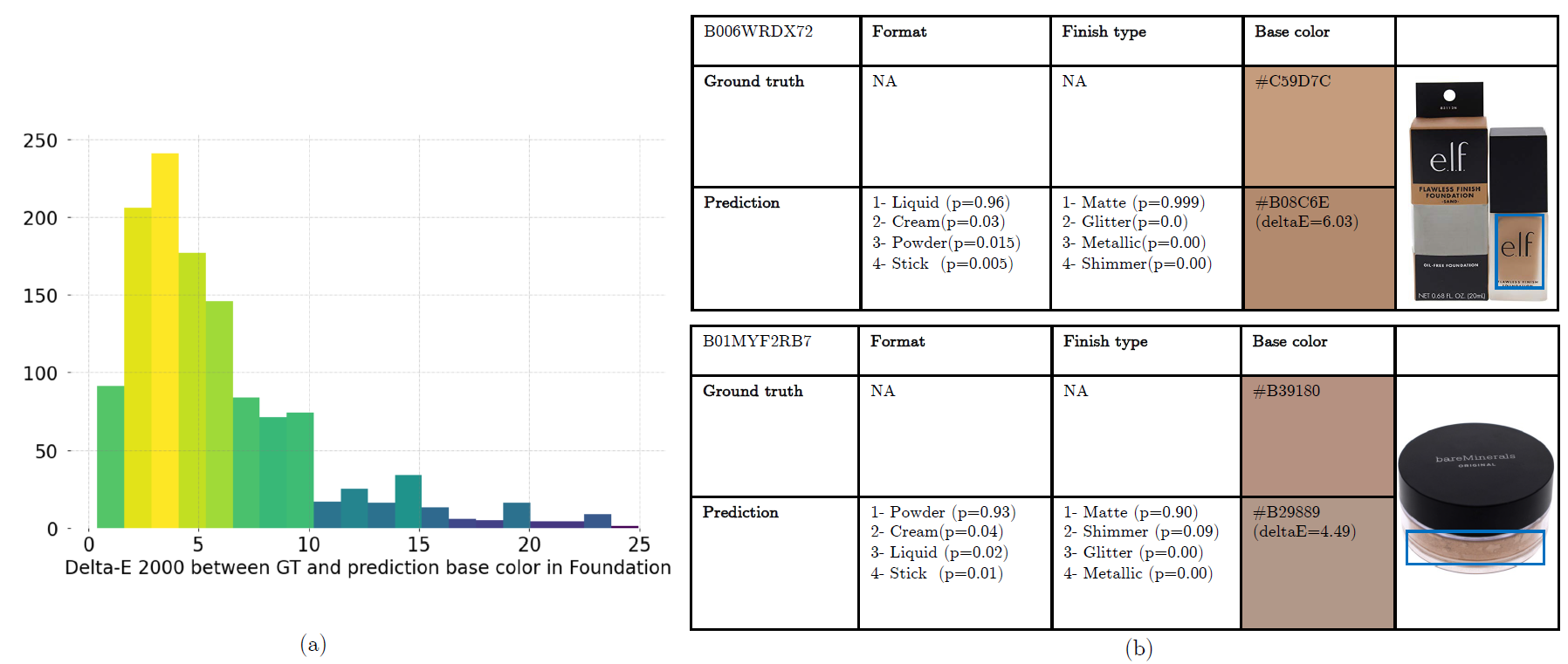}
  \caption{Performance visualization of DeltaE distance histograms for the base color predictions in foundation makeup products along with two sample results.}
   \label{fig:foundation_MPE}
\end{figure}

\begin{figure}[htb]
  \centering
  \includegraphics[width=.9\linewidth]{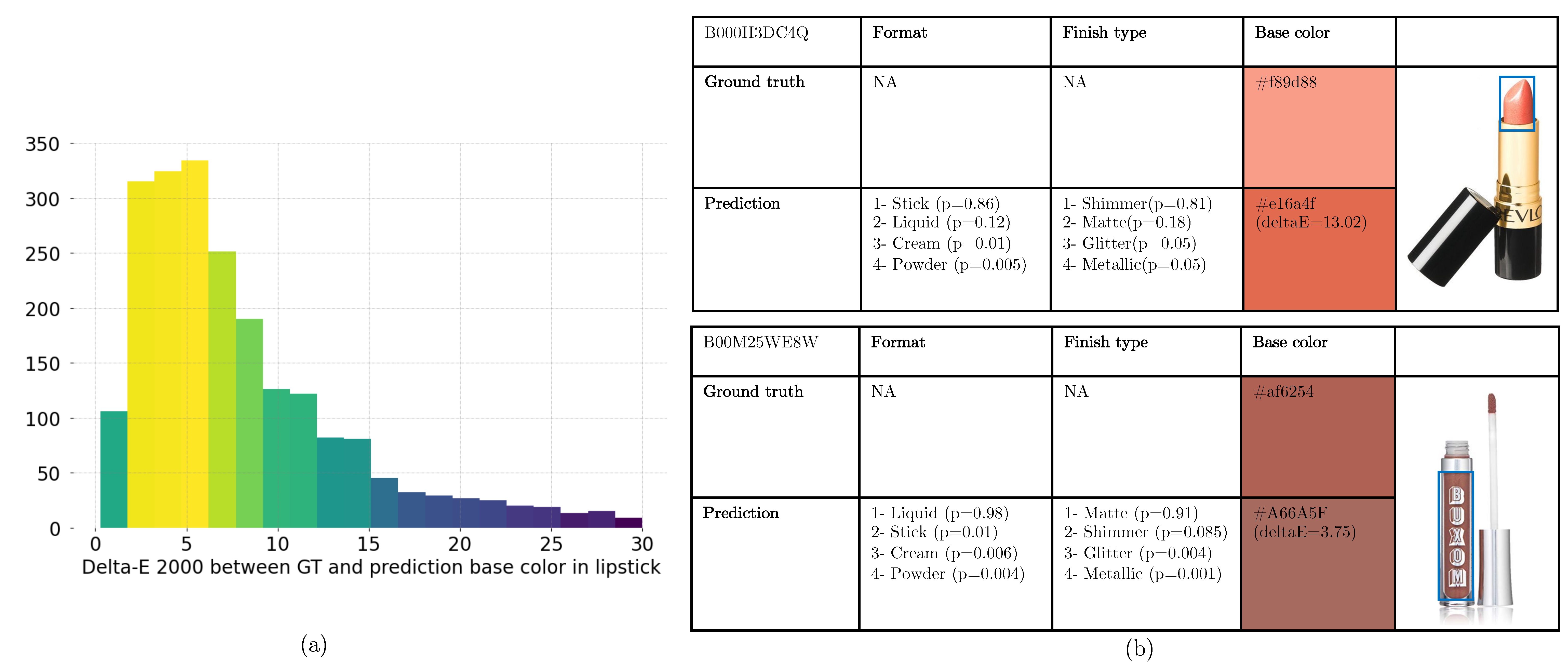}
  \caption{Performance visualization of DeltaE distance histograms for the base color predictions in lipstick makeup products along with two sample results.}
   \label{fig:lipstick_MPE}
\end{figure}

\subsection{Extension to other makeup categories}
The versatility of our proposed method extends beyond eyeshadow alone, as it can be effectively applied to other makeup categories such as lipstick, foundation, and more. To demonstrate its adaptability, we evaluated the performance of our pre-trained model by testing it on datasets from these additional categories, which further validates its broad applicability.

For this experiment, a total of 1,240 "foundation" product pages were crawled from online shopping websites. Each of these products was meticulously annotated for color information, ensuring a comprehensive dataset for our analysis. The model demonstrated promising performance, with an average deltaE score of 5.86. In Fig~\ref{fig:foundation_MPE}.(a), we present histograms illustrating the distribution of deltaE scores between the ground truth and our color models' predictions. The data analysis reveals that the majority of the deltaE scores for the base color models are below 10.  Fig~\ref{fig:foundation_MPE}.(b) showcases two representative samples of foundation products alongside their corresponding model predictions and ground truth.


In another experiment, we gathered a dataset of 2,165 "lipstick" products. This dataset was carefully annotated, and the experiment was conducted following the same procedure as mentioned earlier. The average deltaE score obtained for this experiment was 7.91. The distribution of deltaE scores is visualized in Fig~\ref{fig:lipstick_MPE}.(a), and sample results are presented in Fig~\ref{fig:lipstick_MPE}.(b). Notably, a deltaE score below 5 indicates that two colors are perceived as highly similar by human observers.

\subsection{Impact in Product Recommendations}

\begin{figure*}[tbh]
  \centering
  \includegraphics[width=0.9\textwidth]{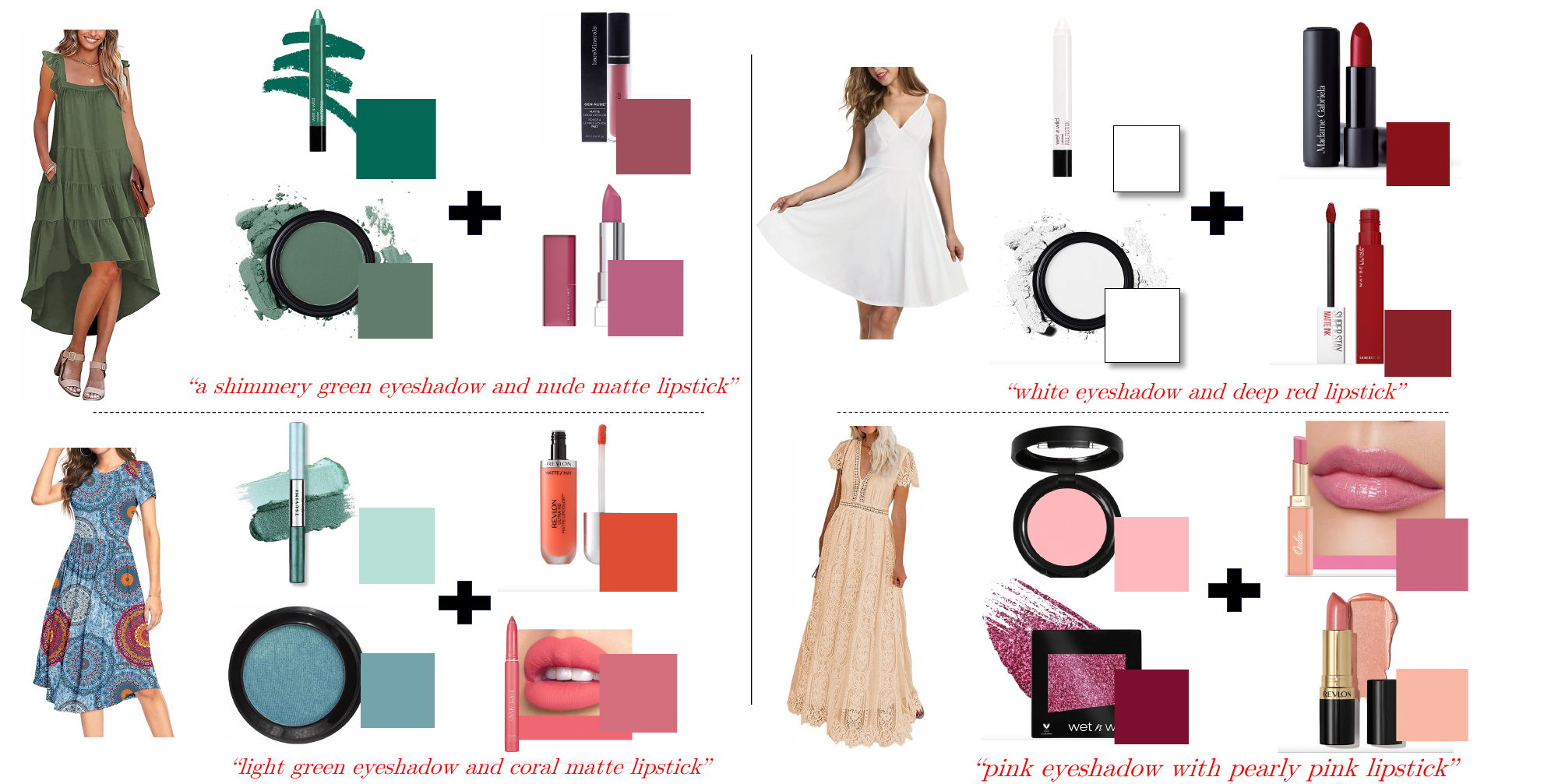}
  \caption{Recommendation makeup products for various clothing dresses. The text on each item is the makeup recommendation for each of the dresses by our in-house makeup professional.}
   \label{fig:reco_fashion}
\end{figure*}

\begin{figure*}[tbh]
  \centering
  \includegraphics[width=0.8\textwidth]{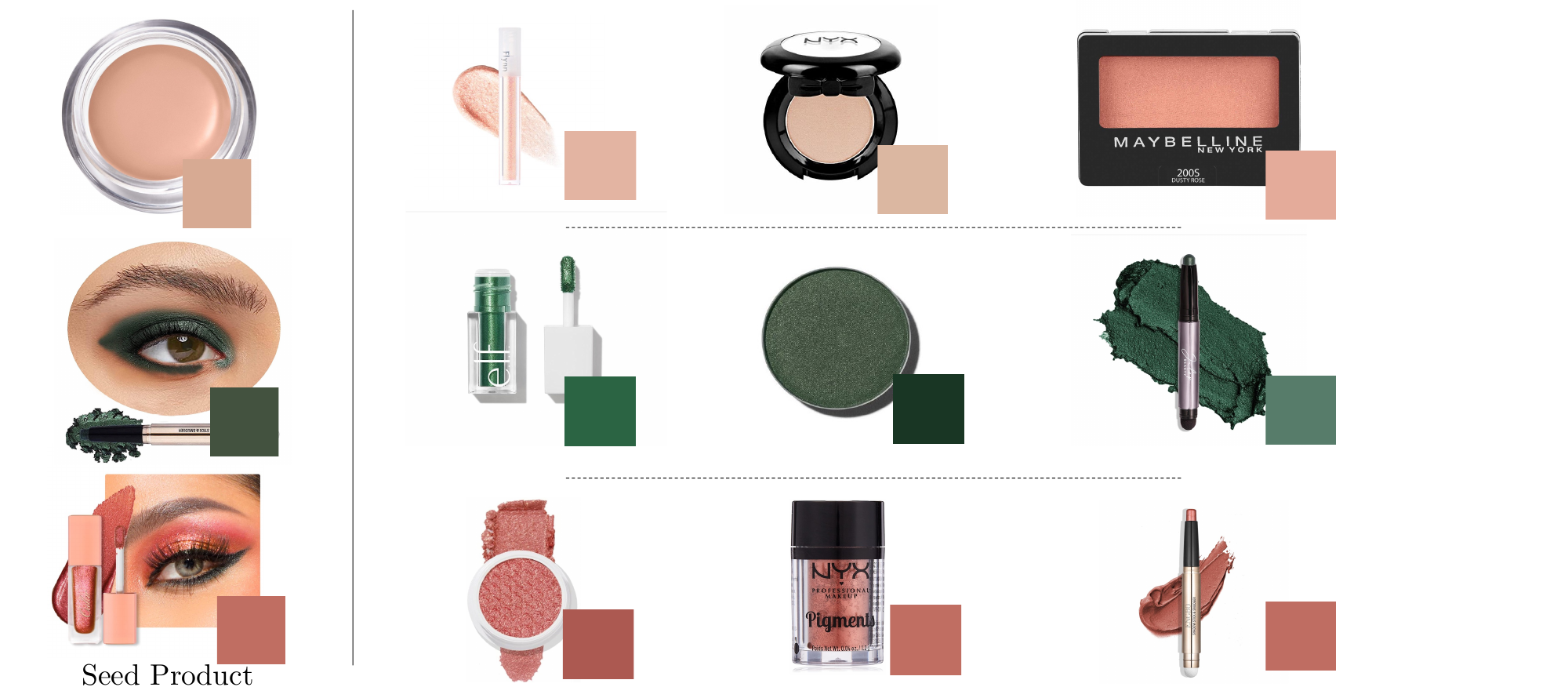}
  \caption{Recommended eyeshadow products on basis of color similarity}
   \label{fig:reco_eyes}
\end{figure*}

Figure~\ref{fig:reco_fashion} showcases the effectiveness of our proposed Makeup Match Maker in providing complementary makeup recommendations for clothing items. The Makeup Match Maker demonstrates its versatility by successfully accommodating various clothing varieties and different types of eyeshadow and lipstick products, all based on the extracted material properties.
In Fig.~\ref{fig:reco_eyes}, we demonstrate the application of similar material properties for eyeshadow products, enabling intra-product recommendations. For our experiments, we set a deltaE color distance limit of $10$ and rank the closest color products to demonstrate the effectiveness of our approach.

\begin{figure*}[tbh]
  \centering
  \includegraphics[width=0.85\textwidth]{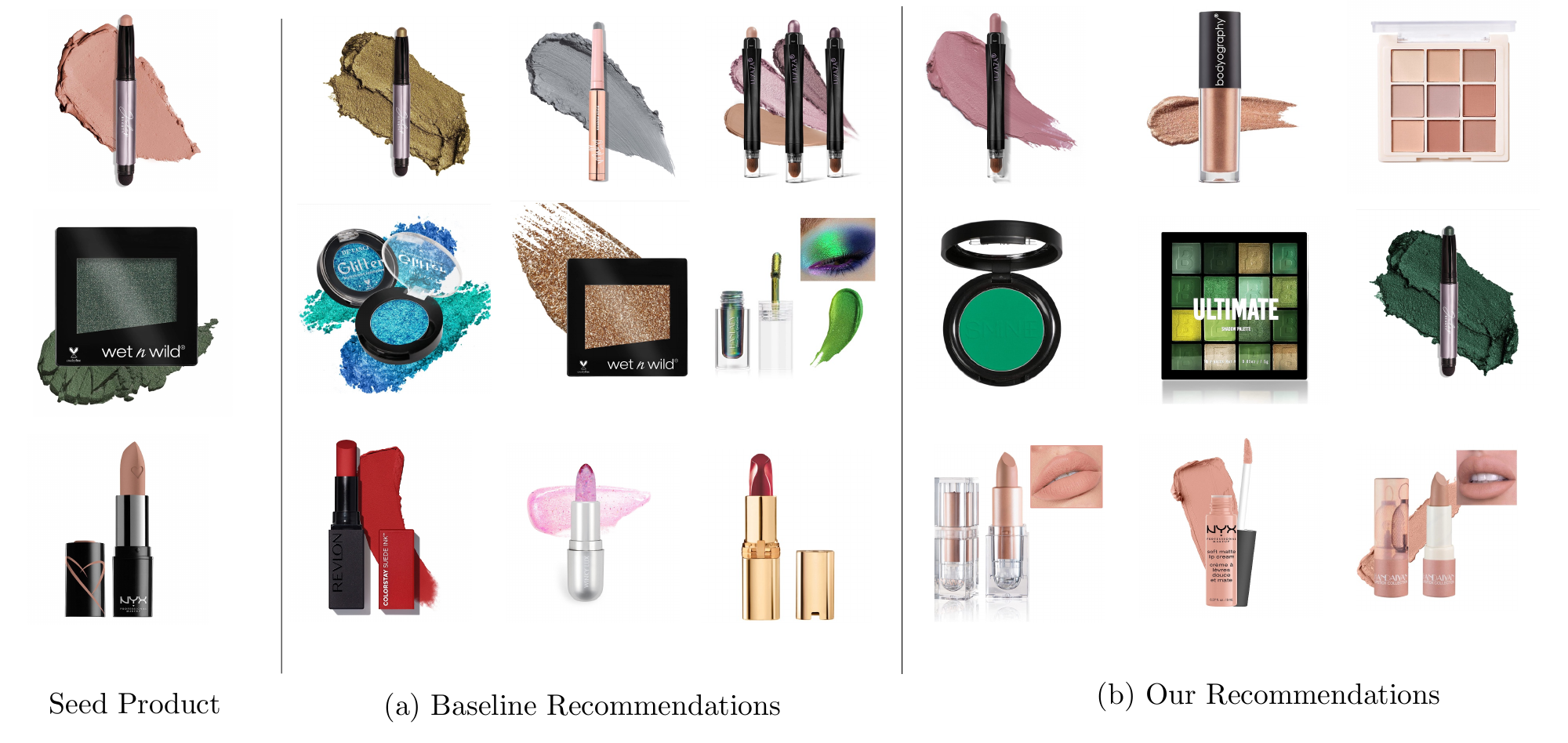}
  \caption{Recommendations for makeup products. (A) shows the baseline recommendations which don't include our material attributes and (B) shows recommendations which have material attributes incorporated  }
   \label{fig:reco_baseline}
\end{figure*}

Additionally, Fig.~\ref{fig:reco_baseline} provides a qualitative demonstration of how the incorporation of our material attributes influences the recommendations. The baseline method primarily relies on brand and product form for recommendations. However, our approach significantly enhances the relevance of the recommendations, as observed in the displayed results.

\begin{figure*}
  \centering
  \includegraphics[width=0.85\textwidth]{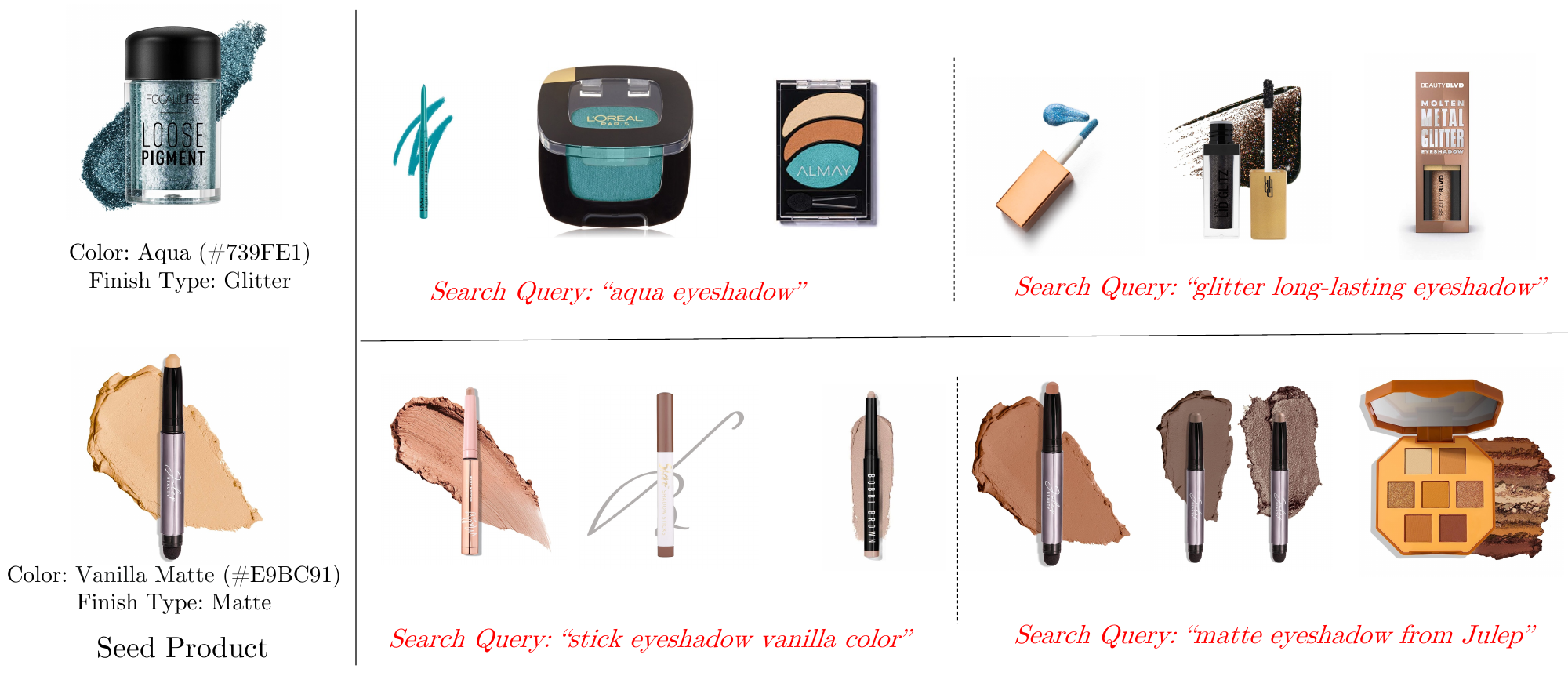}
  \caption{Recommendations for eyeshadow products depending on search query by customer. On the left we have the product and on the right we show the recommended products given the initial search query.  }
   \label{fig:reco_search}
\end{figure*}

In real-world applications, our approach can be combined with other structured attributes and ranked based on user shopping history. Figure~\ref{fig:reco_search} demonstrates how we incorporate extracted properties with available attributes like brand or product type to enhance recommendations. For instance, if a user's initial search query mentions a stick eyeshadow and a specific color preference, our recommendation system intelligently combines these attributes to suggest products that encompass both criteria. This exemplifies our ultimate objective of leveraging these crucial properties to impact recommendations and align them more closely with users' intent when searching for makeup products.
To evaluate the effectiveness of our recommendation algorithm, we conducted an internal user study with 60 participants from diverse backgrounds. Each participant was presented with a combination of recommended products generated using both the baseline and our updated strategy. The task assigned to the participants was to select the item they were most likely to click on, based on the current item. During the study, we evaluated 20 products for both lipstick and eyeshadow with each participant. The results demonstrated that 78\% of the participants chose to click on products recommended by our strategy. This outcome from the user study provides robust evidence for the significance of material attributes in makeup recommendations. It reinforces the notion that incorporating these attributes substantially improves the relevance and appeal of our recommendations to users.

\section{Conclusion}

We proposed a novel learning-based pipeline that utilizes multiple neural network models to extract material properties from makeup products. Our comprehensive pipeline encompasses various models dedicated to image selection, format classification, shade region detection, base/reflective color estimation, and finish type classification. This approach allows for accurate material property extraction from eyeshadow products. Furthermore, we demonstrated the scalability of our model to other makeup categories, such as lipstick and foundation products. Additionally, we showcased the efficacy of utilizing these extracted material properties for successful cross-category product matching.



\bibliographystyle{ACM-Reference-Format}
\bibliography{acmart}

\appendix

\end{document}